\documentclass[sigconf, screen]{acmart}
\usepackage{diagbox}

\AtBeginDocument{%
  \providecommand\BibTeX{{%
    \normalfont B\kern-0.5em{\scshape i\kern-0.25em b}\kern-0.8em\TeX}}}

\copyrightyear{2021} 
\acmYear{2021} 
\setcopyright{acmlicensed}\acmConference[FAccT '21]{Conference on Fairness, Accountability, and Transparency}{March 3--10, 2021}{Virtual Event, Canada}
\acmBooktitle{Conference on Fairness, Accountability, and Transparency (FAccT '21), March 3--10, 2021, Virtual Event, Canada}
\acmPrice{15.00}
\acmDOI{10.1145/3442188.3445920}
\acmISBN{978-1-4503-8309-7/21/03}

\acmYear{2021}\copyrightyear{2021}
\setcopyright{acmlicensed}
\acmConference[FAccT '21]{ACM Conference on Fairness, Accountability, and Transparency}{March 3--10, 2021}{Virtual Event, Canada}
\acmBooktitle{ACM Conference on Fairness, Accountability, and Transparency (FAccT '21), March 3--10, 2021, Virtual Event, Canada}
\acmPrice{15.00}
\acmDOI{10.1145/3442188.3445920}
\acmISBN{978-1-4503-8309-7/21/03}

\begin{document}

\title{One Label, One Billion Faces: Usage and Consistency of Racial Categories in Computer Vision}







\author{Zaid Khan}
\affiliation{\institution{Northeastern University}}
\email{khan.za@northeastern.edu}

\author{Yun Fu}
\affiliation{\institution{Northeastern University}}
\email{yun.fu@ece.neu.edu}


\begin{abstract}
Computer vision is widely deployed, has highly visible, society-altering applications, and documented problems with bias and representation. Datasets are critical for benchmarking progress in fair computer vision, and often employ broad racial categories as population groups for measuring group fairness. Similarly, diversity is often measured in computer vision datasets by ascribing and counting categorical race labels. However, racial categories are ill-defined, unstable temporally and geographically, and have a problematic history of scientific use. 
Although the racial categories used across datasets are superficially similar, the complexity of human race perception suggests the racial system encoded by one dataset may be substantially inconsistent with another. Using the insight that a classifier can learn the racial system encoded by a dataset, we conduct an empirical study of computer vision datasets supplying categorical race labels for face images to determine the cross-dataset consistency and generalization of racial categories. We find that each dataset encodes a substantially unique racial system, despite nominally equivalent racial categories, and some racial categories are systemically less consistent than others across datasets. 
We find evidence that racial categories encode stereotypes, and exclude ethnic groups from categories on the basis of nonconformity to stereotypes.
Representing a billion humans under one racial category may obscure disparities and create new ones by encoding stereotypes of racial systems. The difficulty of adequately converting the abstract concept of race into a tool for measuring fairness underscores the need for a method more flexible and culturally aware than racial categories.  
\end{abstract}




\maketitle

\section{Introduction}

\begin{figure*}
  \centering
  \includegraphics[width=\textwidth]{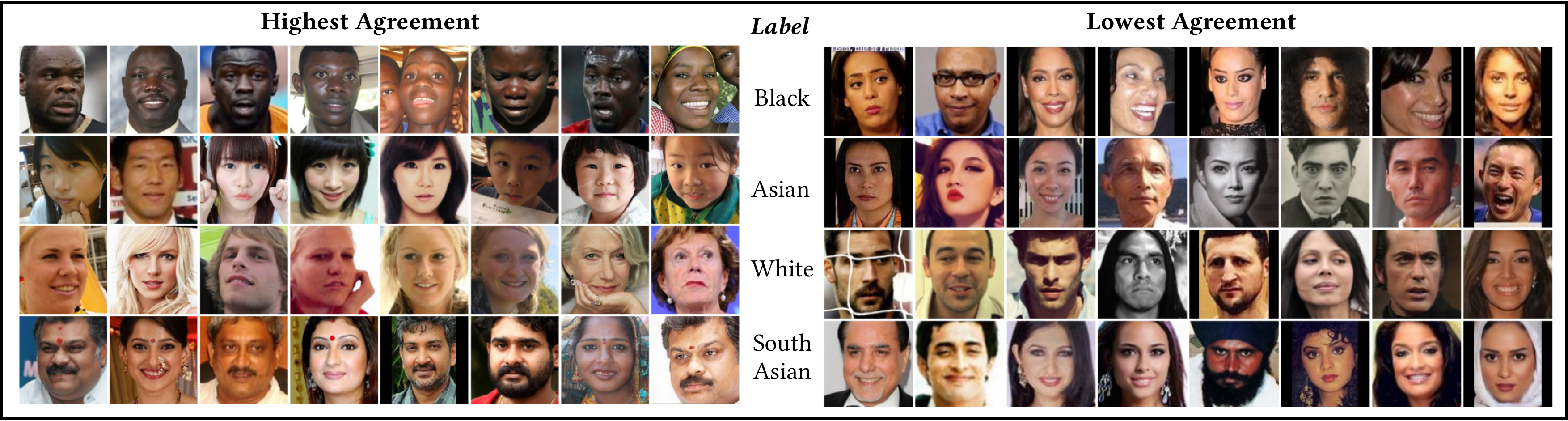}
  \caption{Classifiers trained on fair computer vision datasets agree strongly about the race of some individuals, while disagreeing about others. Racial stereotypes can be seen in the patterns of agreement and disagreement. For example, classifiers were most likely to agree about the race of a individual labeled White by a dataset annotator if the individual was blond.}
  \Description{The 1907 Franklin Model D roadster.}
\end{figure*}
Datasets are a primary driver of progress in machine learning and computer vision, and each dataset reflects human values. Many of the most visible applications of computer vision require datasets consisting of human faces: face recognition\cite{dingGenerativeOneShotFace2019}, kinship\cite{robinsonFamiliesWildMultimedia2020}, demographic estimation \cite{fuLearningRaceFace2014}, emotion recognition \cite{zafeiriouAffWildValenceArousal2017}, and generative modeling \cite{karrasProgressiveGrowingGANs2018}. The datasets driving face-centric computer vision often come with racial annotations of identity, expressed as a race category assigned to each face. Racial categories are also prevalent in the field of fair computer vision, which aims to increase the conformity of computer vision systems to human notions and ideas of fairness. However, little attention is given to the validity, construction, or stability of these categories despite their contentious origins and problematic history of scientific use.

Although these categories are given equivalent names across datasets (e.g., Black, White, etc), \emph{how equivalent are they?} If different datasets have substantially different concepts of race, progress in reducing bias against a group as measured on one dataset is unlikely to transfer to another. Extreme agreement between the datasets could be likewise troubling: race is an abstract, fuzzy notion, and highly consistent representations of a racial group across datasets could be indicative of stereotyping or reification of the physical characteristics of race. Our work is the first to empirically study category bias, consistency, and generalization in the context of racial categories and fair computer vision. Specifically, we quantify the degree to which fair computer vision datasets agree on what a racial category consists of, who belongs to it, how strongly they belong to it, and how consistent these categories are across datasets.

It is tempting to believe fairness can be purely mathematical and independent of the categories used to construct groups, but measuring the fairness of systems \emph{in practice}, or understanding the impact of computer vision in relation to the physical world, necessarily requires references to groups which exist in the real world, however loosely. Datasets attempt to encode these social groupings - commonly racial categories - through an assignment of racial labels to faces so that group fairness \cite{vermaFairnessDefinitionsExplained2018} can be measured. Group fairness aims to equalize a metric such as accuracy across groups defined by a protected characteristic, such as racial categories.

Our work is related in spirit to Hanna et.al, \cite{hannaCriticalRaceMethodology2019} as well as Benthall and Haynes \cite{benthallRacialCategoriesMachine2019}, who both  critique the usage of racial categories from the perspective of critical race theory. We draw on Scheurman et.al's \cite{scheuermanHowWeVe2020} survey of identity classification schemes, definitions, and annotation methods in computer vision. Our work is heavily inspired by Torrallba and Efros \cite{torralbaUnbiasedLookDataset2011}, who similarly use an ensemble of classifiers to empirically study category bias and generalization in the context of general object detection and image classification datasets.

We identify the emergence of a standard racial classification scheme, and use four fair computer vision datasets which implement the scheme to study the consistency of the racial categories and the bias of fair computer vision datasets. We hand-collect an additional, small scale dataset to study how ethnic groups fit into the stereotypes of racial categories encoded into datasets. Our contributions can be summarized as follows: 
\begin{enumerate}
    \item We develop a method to empirically study the generalization and consistency of racial categories across datasets. 
    \item We quantify the generalization and consistency of the racial systems encoded in four commonly used fair CV datasets.
    \item We study how strongly each racial category is represented by a stereotypical presentation unique to each dataset, and who those stereotypes include and exclude.
    \item We determine the degree to which the racial systems in fair computer vision datasets are self-consistent.
    \item We investigate how strongly individuals in biometric datasets are ascribed a racial category, and find systematic differences. 
\end{enumerate}

Every facial image dataset that uses racial categories encodes a racial system - a set of rules and heuristics for assigning racial categories based on facial features. We use a classifier trained on the faces and racial labels in a dataset as an approximation of the racial system of the dataset. By training a classifier for each dataset, we obtain an ensemble of classifiers, each of which has learned the racial system of the dataset it was trained on. By asking each classifier in the ensemble to predict a racial category for the same face, the racial systems of each dataset can be directly interrogated and compared. We use such an ensemble as the basis for our experiments.

The results are intriguing, and we argue that they show racial categories may simply be unreliable as indicators of identity. We find that some racial categories are much more consistently defined than others, and datasets seem to systematically encode stereotypes of racial categories. Our work also opens up concerns about the ability of these racial categories to represent the diversity of human faces. We show that some ethnic groups are excluded from the racial category that they "should" belong to given the straightforward interpretation of those racial categories. We critique and highlight further problems with racial categories in the context of cross-cultural validity and impact on ethnic minorities.

\section{Related Literature}
\subsection{Critical Appraisals of Race in ML}
Historically, the computer vision community has not engaged with critical perspectives on race, typically treating racial categories as commonsense concepts. Benthall and Haynes \cite{benthallRacialCategoriesMachine2019} highlight the instability of racial categories and argue that racial categories have been historically used for stigmatizing individuals, and using racial categories for fairness risks reifying those racial categories and reproducing inequity. In \cite{hannaCriticalRaceMethodology2019}, Hanna, et.al call attention to the multidimensionality of race and the need to critically assess and justify the choice of racial categories and schemes. Monea \cite{moneaRaceComputerVision} catalogues racially problematic incidents in computer vision history and highlights the hypervisbility and reliance on skin color in racial classification schemes.

\subsection{Dataset Audits}
The literature on dataset bias reveals problems endemic to machine learning as a whole, which are only magnified by the complexities of racial identity. Our approach and exposition is heavily inspired by \cite{torralbaUnbiasedLookDataset2011}, who used an ensemble of classifiers to study dataset bias and cross-dataset generalization in commonly used datasets for object detection and classification, noting substantial differences in the representation of nominally similar categories across datasets. We draw on the survey of computer vision datasets in \cite{scheuermanHowWeVe2020}, who enumerate the diverse and ill-defined array of racial classification schemes and definitions in computer vision datasets. Geiger et.al \cite{geigerGarbageGarbageOut2019} find significant variability in the quality of human annotations in a case study of papers on tweet classification. 

\subsection{Racial Classification, Ancestry, Genetics}
Recent advances in population genetics have allowed researchers to interrogate the validity of race from the perspective of genetics and ethnicity. The science of ancient DNA has shown that most modern day human populations are relatively recent mixtures of more ancient populations \cite{lazaridisAncientHumanGenomes2014} who do not fit into modern day racial categories. In addition, contemporary human populations have been shaped by repeated admixture events \cite{hellenthalGeneticAtlasHuman2014} resulting in large-scale gene flow between populations widely separated by geography. Although there are geographically structured difference in genes and correlations between traditional concepts of race and geography, there is substantial genetic overlap between traditional racial groups \cite{jordeGeneticVariationClassification2004}, invalidating the treatment of races as discrete, disjoint groups. 

\subsection{Soft Biometrics}
The field of soft biometrics encompasses notions of identity and traits which are not unique to an individual, but carry information meaningful to society for identification. Canonical examples of soft biometrics are kinship \cite{robinsonFamiliesWildMultimedia2020}, age \cite{chenDeepAgeEstimation2018}, gender \cite{ozbulakHowTransferableAre2016}, and demographic information \cite{fuLearningRaceFace2014}. The line of work most relevant to our work is coarse-grained demographic estimation, which aims to predict a broad racial category for a face. One of the primary challenges in soft biometrics is that \emph{observed traits} may not reflect the ground truth of identity: for example, a person can look much younger than they are, or two individuals may look like sisters due to similarity in appearance. 

\subsection{Cognitive \& Social Perspectives}
Despite the treatment of race as a commonsense, natural kind, research has shown that specialized race perception is unlikely to be an evolved trait or a byproduct of reasoning about natural kinds \cite{cosmidesPerceptionsRace2003}. Edgar et.al \cite{edgarInterObserverAgreementSubjects2011} demonstrate that race estimations in medical records are less consistent for some racial categories than others. Pirtle and Brown \cite{lasterpirtleInconsistencyExpressedObserved2016} draw attention to the mental health consequences of inconsistent racial identification on individuals, and pinpoint Hispanic and Indigenous Americans as two groups who are often perceived inconsistently relative to African and European Americans.

\subsection{Algorithmic Fairness \& Datasets}
Group fairness \cite{vermaFairnessDefinitionsExplained2018} is one of the most easily understood formal notions of fairness, and maps neatly to legal notions of fairness by aiming to equalize a metric across groups, which are often defined by protected characteristics such as race. A vibrant ecosystem of datasets has sprung up to support evaluations of group fairness. Some of most prominent of these are datasets which supply categorical race labels, such as FairFace \cite{karkkainenFairFaceFaceAttribute2019}, BFW \cite{robinson2020facebias}, RFW \cite{wangRacialFacesWild2019}, and LAOFIW \cite{alviTurningBlindEye2019} using compatible racial categories. Other datasets, such as PPB \cite{buolamwiniGenderShadesIntersectional}  and Diversity in Faces \cite{merlerDiversityFaces2019}, use phenotype annotations in lieu of racial categories. Audits of algorithms, APIs, and software  \cite{phillipsOtherRaceEffectFace},\cite{grotherFaceRecognitionVendor2019},\cite{krishnapriyaCharacterizingVariabilityFace2019} have found statistically significant disparities in performance across groups and have been critical to bringing racial disparities to light.

\section{Racial Categories}
\subsection{Usage in Fair Computer Vision}
Racial categories are used without being defined or only loosely and nebulously defined \cite{scheuermanHowWeVe2020} - researchers either take the existence and definitions of racial categories as a given that does not have to be justified, or adopt a "I know it when I see it" approach to racial categories. Given that the categories are allusions to both geographic origin and physical characteristics, it is understandable that deeper discussion of these categories is avoided because it veers into unpleasant territory. This leads to myriad systems of racial classifications and terminology - some of debatable coherence, such as grouping together "people with ancestral origins in Sub-Saharan Africa, India, Bangladesh, Bhutan, among others" \cite{moralesSensitiveNetsLearningAgnostic2020}, and others which could be considered offensive, such as "Mongoloid".

Racial categories are used to define groups of interest in fair machine learning because these categories are believed to reflect social groups in the real world. While Benthall and Haynes \cite{benthallRacialCategoriesMachine2019} highlight issues with the usage of racial categories themselves - recommending their replacement - Hanna et.al \cite{hannaCriticalRaceMethodology2019} argue that racial categories cannot be ignored, as individuals are affected by the racial category others \emph{believe} them to be. 

The construction of large-scale, fair computer vision datasets typically involves collecting a large number of face images, either from an established dataset or web scraping, and then ascribing a racial category to each of the collected faces. Scheurman et.al., \cite{scheuermanHowWeVe2020} detail the myriad racial categories and classification schemes used in computer vision, but we identify and focus on a commonly used set of racial categories: East Asian, South Asian, White, and Black. Occasionally, these categories are referred to by different names - (e.g., Indian instead of South Asian). These categories are partially based on US Census categories and visual distinctness of the categories \cite{karkkainenFairFaceFaceAttribute2019} and believed to "encompass enough of the global population to be useful" \cite{fuLearningRaceFace2014}. 

Crucially, the annotation processes for computer vision datasets largely involve a "moment of identification" by the annotators \cite{scheuermanHowWeVe2020}, in which the annotators ascribe a racial label onto an individual based on their physical characteristics. This is different from data collection regimes in other domains, where racial labels may be provided by individuals themselves. However, the annotation of observed race in computer vision datasets provides a unique opportunity to study the social construction of race and race perception.

\subsection{Specific Problems with Racial Categories}
We identify four specific problems with the usage of racial categories. First, racial categories are ill-defined, arbitrary and implicitly tied loosely to geographic origin. Second, given that racial categories are implicitly references to geographic origin, their extremely broad, continent-spanning construction would result in individuals with drastically different appearances and ethnic identities being grouped incongruously into the same racial category if the racial categories were interpreted literally. Thus racial categories must be understood both as references to geographic origin as well as physical characteristics. Third, racial categories conspicuously erase differences between ethnic groups and group billions of people into one category. Fourth, a system of identity in which a face is assigned a singular racial category is incapable of expressing a substantial proportion of human diversity and variation.

\subsubsection{Racial Categories Are Badly Defined}
The Indian/South Asian category presents an excellent example of the pitfalls of racial categories. If the Indian category refers only to the \emph{country} of India, it is obviously arbitrary - the borders of India represent the partitioning of a colonial empire on political grounds. Therefore, it is reasonable to take the Indian category as representing the entirety of the Indian subcontinent, or South Asia. This presents another set of problems, as expanding the region we consider the "home" of a racial category also expands our borders, and introduces another, larger set of shared borders along which populations should be visually indistinguishable due to proximity, admixture, and shared culture.
\subsubsection{Naive Interpretations of Racial Categories Are Incongruous}
Because racial categories are defined with respect to, and largely correspond with geographic regions, they include populations with a wide range of phenotypical variation, from those most steretotypical of a racial category to  racially ambiguous individuals or those who may appear to be of a different observed race altogether. As an example, the South Asian racial category should include populations in Northeast India, who may exhibit traits which are more common in East Asia. However, including individuals which appear East Asian in a racial category distinct from East Asian would violate the common, lay understanding of racial categories.  
\subsubsection{Racial Categories Erase Ethnic Differences and Identity}
Racial categories are continent-spanning groupings. Every racial category subsumes hundreds of ethnic groups with distinct languages, cultures, separation in space and time, and phenotypes. Ethnic groups even span racial lines \cite{obachDemonstratingSocialConstruction1999} and racial categories can fractionalize ethnic groups, placing some members in one racial category, and others in a different category. Evaluations of fairness which only use racial categories cannot quantify bias against individual ethnic groups, and ascribing a privileged identity to a member of an underprivileged group could result in individual harm by algorithmic intervention.

\subsubsection{Racial Categories Aren't Expressive Enough}
The often employed, standard set of racial categories (Asian, Black, White, South Asian) which we study in this paper is, at a glance, incapable of representing a substantial number of humans. It obviously excludes indigenous peoples of the Americas, and it is unclear where the hundreds of millions of people who live in the Near East, Middle East, or North Africa should be placed. One can consider extending the number of racial categories used, but racial categories will always be incapable of expressing multiracial individuals, or racially ambiguous individuals. National origin or ethnic origin can be utilized, but the borders of countries are often the results of historical circumstance and don't reflect differences in appearance, and many countries are not racially homogeneous. 

\subsection{Reasons to Expect Inconsistency}
The four datasets we consider all have racial categories which are nominally equivalent: East Asian, South Asian, White, and Black. We chose to exclude datasets from our analysis that use obviously different racial categories , such as DemogPairs \cite{hupontDemogPairsQuantifyingImpact2019b}. We expect to find inconsistency between these nominally equivalent racial categories for several reasons. First, despite the commonsense belief in racial categories, there are differences in how individuals perceive race. Second, the huge population size of racial categories means that samples in datasets drawn from the racial categories may represent drastically different faces due to the difference between sample size and population size.

 \subsubsection{Inconsistent Human Race Perception} The self-identification of humans affects \cite{felicianoShadesRaceHow2016} \cite{paukerMultiracialFacesHow2009} how they classify the race of others. In addition, human observers are notably less congruent when identifying \cite{hermanYouSeeWhat2010}\cite{lasterpirtleInconsistencyExpressedObserved2016} faces from non-White or non-Black backgrounds. Telles \cite{tellesRacialAmbiguityBrazilian2002} found that observers were more consistent when assessing the race of lighter-skinned individuals than darker skinned individuals. Taken together, these facts paint a picture of a human annotator whose racial system for assigning race to a face depends substantially on the background of the annotator and is more variable for individuals with some characteristics than others.

\subsubsection{Large, Heterogeneous Populations} Even if all annotators had the same set of rules for mapping facial traits to racial categories, each dataset must represent racial categories consisting of several billion people with tens of thousands of faces. The small size of these samples relative to the heterogeneity, diversity, and size of populations being sampled implies that each dataset may contain faces drawn from parts of the population which were not sampled in other race datasets. Thus, even in the unlikely case that annotators subscribe to the same racial system, nominally similar racial categories may contain different faces.

\section{Experiments}
Our approach is based heavily on Torralba and Efros \cite{torralbaUnbiasedLookDataset2011}, who use an ensemble of classifiers to study dataset bias by testing how well nominally equivalent categories generalize from one dataset to the next. We broadly follow Torralba and Efros's approach, though we focus on \emph{consistency} between classifiers as well as accuracy against the "true" labels of a dataset. The key insight of this approach is that a sufficiently powerful classifier trained on a dataset will capture the heuristics and rules used by the dataset creators or annotators to assign a class to an image, and these classifiers can then be compared to quantify the degree of difference between the classification systems in a dataset.

In the first experiment, we train the ensemble of classifiers and measure cross dataset generalization.
In the second experiment, we split each dataset into 3 equal-sized, disjoint subsets and train a new classifier on each disjoint part to gain insight into the self-consistency of each dataset and the effect of dataset size on the ability to represent a racial category. 
In the third experiment, we use the predictions from the classifier ensemble to investigate how strongly individuals with multiple pictures are associated to a racial category
In the fourth we study category bias in datasets from a different perspective by using a dataset of ethnic groups to examine stereotypes learnt by the ensemble of classifiers. 

\begin{table*}[]
\caption{Cross-Dataset Generalization of Racial Categories}
\label{tab:x-ds-generalization}
\centering
\begin{tabular}{@{}lccccccccc@{}}
\toprule
& \multicolumn{2}{c}{Source Dataset}  & \multicolumn{4}{c}{Target Dataset} & \multicolumn{3}{c}{Generalization} \\ 
\cmidrule(l{0.5em}r{0.5em}){2-3} \cmidrule(l{0.5em}r{0.5em}){4-7} \cmidrule(l{0.5em}r{0.5em}){8-10} 
Classifier        & Faces   & Identities & BFW & FairFace  & LAOFIW & RFW & Self & Others & Change \\ 
\cmidrule(r){1-1} \cmidrule[0.6pt](l{0.5em}){2-10}
BFW     &  20000 & 800 & 0.903 & 0.641    & 0.714  & 0.811  & 0.903  & 0.722  & -20.0\%        \\
FairFace &  71781 & 71781 & 0.779 & 0.857    & 0.782  & 0.870  & 0.857  & 0.810  & -5.48\%      \\
LAOFIW   &  9906  & 9906  & 0.830 & 0.644    & 0.809  & 0.820  & 0.809  & 0.765  & -5.44\%     \\
RFW     &  40573 & 11400 & 0.895 & 0.748    & 0.781  & 0.930  & 0.930  & 0.808  & -13.1\%       \\
\bottomrule
\end{tabular}
\end{table*}
\subsection{Preliminaries}

\subsubsection{Datasets}
\label{sec:dataprep}
We use four fair computer vision datasets: FairFace \cite{karkkainenFairFaceFaceAttribute2019}, BFW \cite{robinson2020facebias}, RFW \cite{wangRacialFacesWild2019}, and LAOFIW \cite{alviTurningBlindEye2019}. These datasets were chosen for several reasons. First, they encode nominally equivalent categories: White, Black, East Asian, and South Asian. Second, FairFace, BFW, and RFW are all built from widely used, large-scale computer vision datasets. FairFace is built from YFCC-100M, BFW is built from VGG2, and RFW from MSCeleb-1M. All four of the datasets have impacted the field of fair ML, either by their use as benchmarks, or by revealing new inequalities between populations. BFW and RFW are \emph{deep} - they have multiple photos per individual, while FairFace and LAOFIW have only one photo per individual, except for incidental duplicates.

Other datasets, such as DivFace \cite{moralesSensitiveNetsLearningAgnostic2020} or DemogPairs \cite{hupontDemogPairsQuantifyingImpact2019b}, use custom, nominally incompatible racial categories and/or mix multiple source datasets. Datasets like 10K US Adult Faces \cite{bainbridgeIntrinsicMemorabilityFace2013} have not been explicitly targeted towards the fair computer vision community and contain image modifications which make them easily distinguishable. Finally, all of the selected datasets are large enough to support the training of a modern CNN for classification. This is not the case for some older datasets, which contain only a few hundred images. 

\subsubsection{Classifier Ensemble}
Our key insight is that an image-based classifier, if trained on the racial categories in a dataset, can learn the system of rules the annotators of the dataset used to assign a racial category to a face. By training multiplier classifiers - one for each dataset - we obtain an ensemble of classifiers, each having learned the racial system encoded by their source dataset. By comparing the predictions of these classifiers, we can quantify how different the racial system in each dataset is. We use a 50-layer ResNet\cite{heDeepResidualLearning2015}, a commonly used backbone for deep vision tasks, as our base classifier. The ResNets use pretrained weights from ImageNet\cite{imagenet_cvpr09}, and are implemented in PyTorch \cite{pytorch_nips_2019}. To set the learning rate, we use the one-cycle policy \cite{smithDisciplinedApproachNeural2018} using the implementation in \cite{howardFastaiLayeredAPI2020}.

\subsubsection{Preprocessing}
To mitigate spurious differences between datasets, we performed a standard preprocessing in face recognition \cite{liuSphereFaceDeepHypersphere2017} by aligning the faces in each dataset using the keypoints detected by an MTCCN \cite{zhangJointFaceDetection2016} so the left eye lies at $(0.35 \times w, 0.35 \times h)$, while simultaneously rotating and rescaling the image to $(w=256, h=256)$ so that the eyes are level and the distance between the eyes is the same in all images, thus ensuring all images have the same dimensions, and contain faces of the same scale. FairFace separates the "Asian" category into "East Asian" and "Southeast Asian" - we combine this into one category. 

\subsubsection{Metrics}
To measure cross-dataset generalization, we use the standard notion of accuracy. To measure the consistency of predictions made by multiple classifiers, we use the Fleiss-$\kappa$ score, a standard measure of inter-annotator agreement \cite{artsteinInterannotatorAgreement2017} applicable to multiple annotators.
Conceptually, the Fleiss-$\kappa$ score is amount of agreement between the annotators beyond that expected by chance. 
Interpretation of the Fleiss-$\kappa$ score can be difficult in absolute terms, so we typically restrict ourselves to comparing Fleiss-$\kappa$ scores relative to one another.

\subsection{Cross-Dataset Generalization}
\label{sec:x-ds-generation}
\subsubsection{Setup}
To understand how well the racial categories from one dataset generalize to another, we construct an ensemble of classifiers, $\mathcal{C} = \{C_{BFW}, C_{FairFace}, C_{RFW}, C_{LAOFIW}\}$. Each classifier is trained on a different source dataset, and all datasets use a common set of a racial categories: $\mathcal{R} = \{ \text{Black}, \text{white}, \text{asian}, \text{indian} \}$, and each face in the dataset comes annotated with a racial category from $\mathcal{R}$. Each classifier is trained to predict a racial category $r \in \mathcal{R}$, given a face. The training set is constructed for each classifier by using 80\% of the faces in the source dataset of the classifier, and using the remaining 20\% as a validation set. As a concrete example, the $C_{FairFace}$ classifier would be trained using 80\% of the faces in FairFace, thus learning the heuristics used by the dataset creators to assign a racial category to a face.

The classifiers were trained with the standard categorical cross-entropy loss. Due to the usage of pretrained weights and a one-cycle learning rate policy \cite{smithDisciplinedApproachNeural2018}, only a small number of iterations were sufficient for model convergence: 40 iterations for RFW and FairFace, and 25 for LAOFIW and BFW. 

\subsubsection{Results} The results are reported in Table \ref{tab:x-ds-generalization} as accuracy scores. All classifiers displayed similar generalization performance, suggesting that there is significant overlap and a core set of rules for assigning racial categories to a face that is shared among all datasets. Notably, final accuracy reached by each classifier on the validation set varies significantly, but the absolute difference between the generalization performance of different datasets is small. In some sense, the easiest racial rulesets were encoded by BFW and RFW, as the BFW and RFW classifiers showed the highest performance on their validation sets. This could be due to the fact that BFW and RFW contain multiple images per identity. 

\subsubsection{Label Noise and Learning} We note that looking only at the accuracy of each classifier to conclude it has insufficiently learned the rules used by the annotators of the dataset is misleading. The classifier only has the cropped faces available, and cannot use contextual information used by the annotator to make racial decisions. For example, research has shown that humans use cues like education level to make determinations of race \cite{tellesRacialAmbiguityBrazilian2002}, which can be inferred from clothing or image context. In the context of the classification problem faced by the model, these phenomena can be interpreted as label noise, and deep learning models are incredibly robust to noisy labels \cite{zhangUnderstandingDeepLearning2017} and can learn meaningful rules even in the presence of high label noise. Furthermore, it is possible that each dataset contains inconsistencies \emph{within itself}. To give a concrete example, the annotators may have assigned a racial label to a face, and then assigned a different racial label to an extremely similar face due to contextual information or human error. We turn next to an investigation of the self-consistency of each dataset. 

\begin{table*}[]
\caption{\label{tab:self-consistency} Self-consistency, reported as Fleiss-$\kappa$ scores.}
\centering
\begin{tabular}{@{}lcccccccc@{}}
\toprule
& \multicolumn{4}{c}{Self-Ensemble Consistency} & \multicolumn{2}{c}{Average Consistencies} & \multicolumn{2}{c}{Percent Changes}\\
 \cmidrule(l{0.5em}r{0.5em}){2-5} \cmidrule(l{0.5em}r{0.5em}){6-7} \cmidrule(l{0.5em}r{0.5em}){8-9}
Target Dataset & BFW           & FairFace & LAOFIW & RFW   & Whole/Mixed & Self-Ensembles & Consistency & Train Set Size \\ \cmidrule(r){1-1} \cmidrule[0.6pt](l{0.5em}){2-9}
BFW            & -             & 0.7      & 0.626  & 0.823 & 0.673  & 0.716             & +6.35\%        & -66\%               \\
FairFace       & 0.615         & -        & 0.485  & 0.689 & 0.549  & 0.596             & +8.62\%        & -66\%               \\
LAOFIW         & 0.754         & 0.819    & -      & 0.754 & 0.709  & 0.776             & +9.4\%         & -66\%               \\
RFW            & 0.774         & 0.828    & 0.636  & -     & 0.727  & 0.746             & +2.61\%        & -66\%    \\
\bottomrule
\end{tabular}
\end{table*}

\subsection{Self-Consistency}
The drop in cross-dataset generalization of racial categories can be explained by three reasons. 
First, different annotators may have conflicting racial systems - they have assigned different racial labels to similar faces, the classifiers have learnt that, and so the classifiers disagree. Second, the rules for assigning race to a face may not be covered by the training data of the classifier. In other words, the face is out of distribution \cite{heinWhyReLUNetworks2019}, and the classifier has never seen a similar face. This is very likely to happen in the case of BFW, which contains only 800 different identities. Third, the rules for assigning racial categories to a face might simply be inconsistent or unlearnable, in which case the classifier would just be memorizing. The reasonably high generalization performance of each classifier indicates this is unlikely to be the case.  We conduct an experiment to provide insight into whether racial systems encoded by the datasets are significantly different and conflicting, or whether the source of the conflict is differences in the data distributions of each dataset.

\subsubsection{Conflicting Racial Systems or Data Distributions?}
A face may be perceived as Asian by one person, but Indian by another. This would correspond to classifiers making conflicting predictions about an individuals race because \emph{the underlying datasets define race differently for that individual}. In the second scenario, the individual the classifiers are asked to classify is not similar to any face present in the classifiers' training datasets. This is possible because the datasets are sampling from populations consisting of billions of individuals, but are limited to a sample size multiple orders of magnitudes smaller than the population. As a result, it is possible that the classifiers only disagree because \emph{the underlying datasets do not contain any faces similar to the face being considered}.

\subsubsection{Setup} To gain insight into whether we are seeing conflicting racial systems or the result of dataset size and facial distribution differences, we devise the following experiment. We create a \emph{self-ensemble} for each dataset by splitting each dataset into 3 disjoint subsets, and training a classifier on each subset. BFW and RFW, contain multiple images per person and are split so that no person appears in multiple splits. Training protocols remain the same. We then evaluate the consistency of each self-ensemble against the ensemble of classifiers trained on the whole datasets 
$\mathcal{C} = \{C_{BFW}, C_{FairFace}, C_{RFW}, C_{LAOFIW}\}$. 

To evaluate the consistency of each self-ensemble, we predict a race label for each face in a target dataset using the self-ensemble. Because each self-ensemble contains 3 classifiers, each trained on a different subset of the source dataset, this results in 3 predictions for every face in the target dataset. We can then obtain the consistency of the self-ensemble predictions using the Fleiss-$\kappa$ score measuring inter-annotator agreement. A similar procedure is followed for the mixed ensemble $\mathcal{C}$ trained on the whole datasets.

\subsubsection{Results} 
The results can be seen in Table \ref{tab:self-consistency}. With the exception of LAOFIW, the predictions of each self-ensemble are more consistent than the predictions of the mixed/whole ensemble $\mathcal{C}$ trained on the whole datasets in Section \ref{sec:x-ds-generation}. Despite a significant drop in the training set size (-66\%), the consistency of the self-ensembles \emph{on average} is higher than the consistency of the mixed/whole ensembles. We interpret this as evidence that arbitrary subsets of each dataset (except LAOFIW), even if much smaller than the whole dataset, are large enough to encode generally consistent rules for assigning racial categories to faces, and suggests that we are seeing conflicts in rules for assigning racial labels to a face rather than disagreements caused by out of distribution problems.

\subsection{Distribution of Racial Disagreements}
\begin{figure*}
  \centering
  \includegraphics[width=\textwidth]{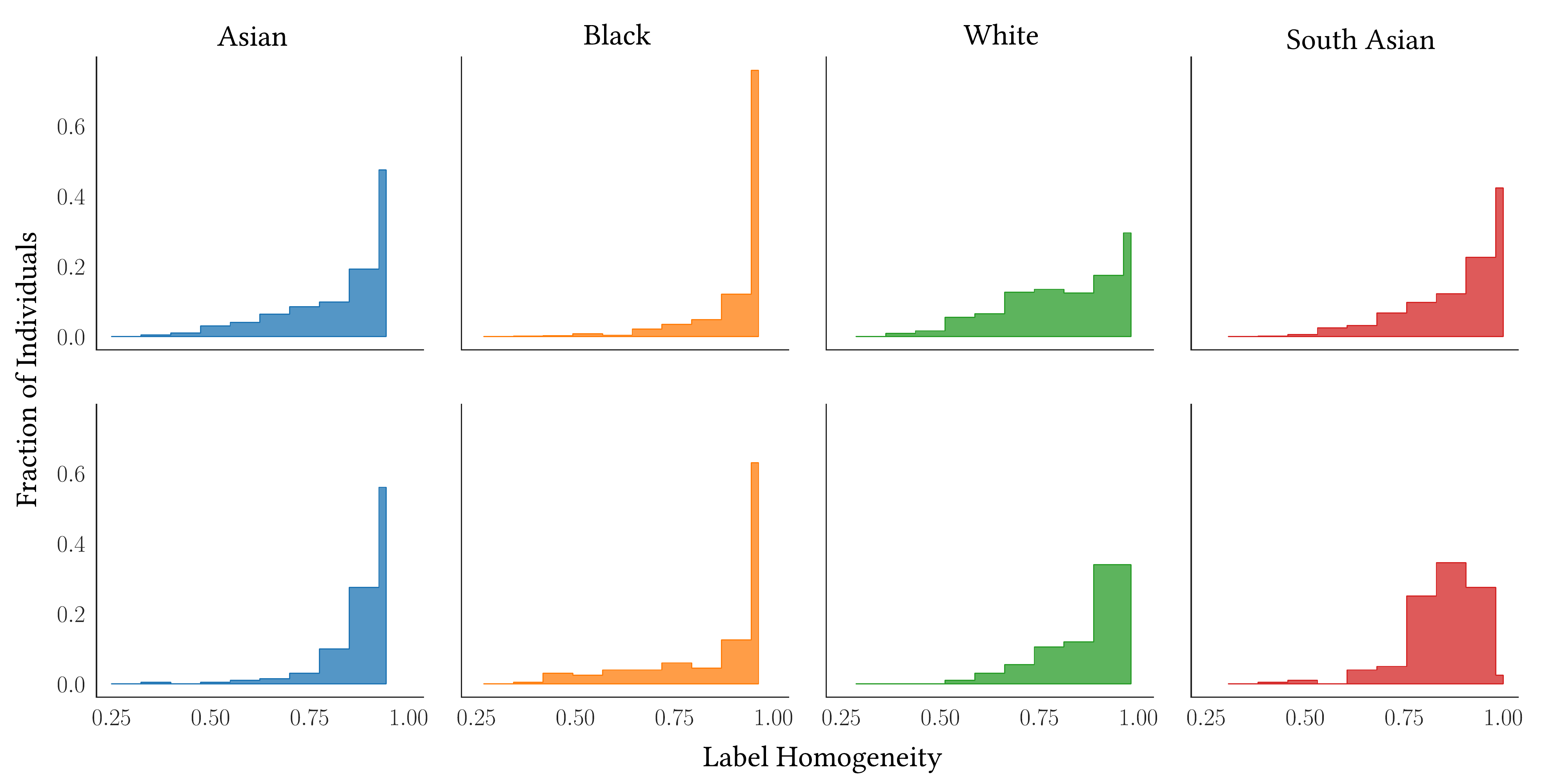}
  \Description{The 1907 Franklin Model D roadster.}
  \caption{Datasets agree strongly on who is Black, as seen by the large fraction of individuals labeled Black who have high homogeneity in their predicted labels. Compare to South Asian or White categories, in which a large fraction of individuals exhibit low homogeneity in their predicted labels, meaning classifiers often disagreed on their racial category.}
  \label{fig:homogeneity}
\end{figure*}

The cross-dataset generalization and self-consistency experiments have shown that racial systems of datasets are self-consistent and inconsistency is caused by conflicting racial systems between datasets. However, the distribution of inconsistency is still unclear. Are some racial categories less consistent and stable across datasets than others, and if so, is the source of the inconsistency largely random variations, or it systematic? We can analyze this question using biometric datasets like BFW and RFW, which contain multiple pictures per individual. If inconsistency is random, we expect to see a large number of individuals who each have a small number of photos in which the classifier ensemble $\mathcal{C}$ disagrees about their race. If inconsistency is systematic, we expect to see a a substantial number of individuals who have a large number of photos in which the classifier ensemble $\mathcal{C}$ disagrees about their race.

\subsubsection{Setup} We use the BFW and RFW datasets, along with the classifier ensemble $\mathcal{C} = \{C_{BFW}, C_{FairFace}, C_{RFW}, C_{LAOFIW}\}$. We obtain a racial category prediction for each face in BFW and RFW. This results in 4 predictions for each image, as there are four classifiers in the classifier ensemble. We then calculate the \emph{label homogeneity} of each individual, which is the fraction of predictions in which the most frequently predicted racial category is present. For example, an individual with 25 photos would receive a total of 100 predictions, 4 for each image. If the most frequently predicted label was white, and white was predicted 90 times, the label homogeneity would be $\frac{90}{100} = 0.9$. Simply put, label homogeneity is another number to describe how strongly the classifiers saw an individual as belonging to a racial category.

\subsubsection{Results} All racial categories, across both BFW and RFW had a substantial number of individuals with very high label homogeneity, meaning the classifier ensemble strongly identified them with the racial category the annotators labeled them as (Figure \ref{fig:homogeneity}). All racial categories also had a substantial number of individuals with low label homogeneity, indicating the classifiers systematically disagreed on the race of the individual, and disagreements are not random. Label homogeneity was much higher for individuals labeled Black than individuals labeled White or South Asian. This is surprising, as there is no a priori reason to expect that identification of a Black individual is easier than that of another racial category. We suspect that datasets either define Blackness very strictly, or have a narrow concept of being Black that is consistent across datasets. 


\begin{table}[]
\caption{Cross-Dataset Consistency of Racial Categories. Reflecting Fig. \ref{fig:homogeneity}, Black is the most consistently defined category across datasets. All numbers are Fleiss-$\kappa$ scores. }
\begin{tabular}{@{}lccccc@{}}
\toprule
Racial Category        & BFW  & FairFace    & LAOFIW     & RFW        & Avg        \\
\cmidrule(r{0.5em}){1-1} \cmidrule(l{0.5em}r{0.5em}){2-5} \cmidrule(l{0.5em}){6-6}
Black    & 0.79 & {\color{blue}0.67}          & 0.76       & {\color{blue}0.9}        & {\color{blue}0.78}       \\
Asian   & {\color{blue}0.81}    & 0.58        & {\color{blue}0.78}       & 0.75       & 0.73       \\
White   & {\color{orange} 0.54} &{\color{orange}0.44}         & 0.75       & 0.73       & {\color{orange}0.615}      \\
South Asian      & 0.73 & 0.6         & {\color{orange}0.62}       & {\color{orange}0.63}       & 0.645      \\
\midrule
Gap         & 50\% & 52.3\%      & 25.8\%     & 42.0\%     & 26.9\% \\
\bottomrule
\end{tabular}
\label{tab:x-ds-cons}
\end{table}

\subsection{Consistency of Racial Categories}
\label{sec:x-ds-consistency}
To understand which racial categories are most stable and least stable across datasets, we measure the consistency of the classifier ensemble $\mathcal{C} = \{C_{BFW}, C_{FairFace}, C_{RFW}, C_{LAOFIW}\}$ per racial category for each dataset. We use the Fleiss-$\kappa$ score as the metric of consistency. 

\subsubsection{Results}
The most stable and consistent category across datasets was Black as seen in Table \ref{tab:x-ds-cons}. The least stable was White, followed closely by South Asian. In general, the Asian and Black categories were much more stable than the South Asian and White categories by a substantial margin. The faces in the Black category of RFW were the most consistently percieved with a Fleiss-$\kappa=0.9$. The faces in the Asian category of BFW were second most consistently percieved, with a Fleiss-$\kappa=0.81$. The faces in the White category of FairFace were the least consistently percieved, with a Fleiss-$\kappa=0.44$, much lower than any other category in any dataset.   

\subsection{Ethnicity and Stereotypes}
The previous experiments paint a picture in which some racial categories are much more consistent across datasets than others. All the racial categories are equally ill-defined, so it is surprising that in practice, datasets are very consistent about who goes in the Black category or the Asian category relative to their inconsistency about the South Asian and White categories. 
We hypothesize that datasets simply include a more diverse set of White and South Asian faces, compared to their limited conception of Asian and Black faces. To investigate this idea, we collect a dataset of ethnic groups. For each geographic region tied to a racial category, we select two ethnic groups originating from that geographic region, taking care to select groups which are significantly separated from each other by geography in the hope that this will maximize differences in phenotypic traits.

\subsubsection{Dataset} We obtain all images from the web by hand to ensure quality. We draw on three sources for images: lists of public figures such as sportspeople and politicians, pictures taken at ethnicity-specific festivals, and images taken by travel photographers in which ethnicity of the subjects is explicitly noted and geolocation data matches the targeted region. We download the images, detect, crop, align, and resize the faces following in the procedure in \ref{sec:dataprep}. 
The ethnicities collected, and the number of images collected, are summarized in Table \ref{tab:ethstats}. 

\subsubsection{Setup} Following the procedure in \ref{sec:x-ds-consistency}, we use the classifier ensemble $\mathcal{C}$ trained in \ref{sec:x-ds-generation}, to predict a racial category for every face in every ethnic group. We then obtain Fleiss-$\kappa$ scores for the ensemble predictions for each ethnic group. This is the same procedure as \ref{sec:x-ds-consistency}, except we group by ethnicity instead of racial category. We also count the most frequently predicted racial category for each ethnic group, which we summarize in Fig. \ref{fig:stereotypes}.

\subsubsection{Results} The most consistently perceived ethnic groups are Iceland and Korean, on which the classifier ensemble $\mathcal{C}$ agreed almost perfectly. The least consistently perceived ethnic group was Balochi, with a Fleiss-$\kappa$ score much lower than the others. The results fit into a clear pattern, visible in Fig. \ref{fig:stereotypes}. The classifier ensemble agrees on the racial categorization of some ethnic groups very strongly, on others, it disagrees very strongly. 
We interpret the appearance of such an effect along ethnic lines to be evidence that racial categories in dataset are partly based on racial stereotypes, and the high level of consistency shown by the classifier ensemble $\mathcal{C}$ on the Black and Asian category is partly a result of all tested datasets containing images for those categories which conform to a common stereotype. 
The Black racial category is the most consistent across all datasets, with a Fleiss-$\kappa=0.78$. However, we argue that this is not because it is easier to identify individuals who are Black, but because datasets contain a narrow conception of who is Black and lack sufficient ethnic diversity. 
When presented with a ethnic group who is likely not racialized adequately due to lack of representation in datasets (Ethiopian), but originates from Africa and could be considered Black by based on geographic origin, the classifier ensemble is very inconsistent.

\begin{table}[]
\caption{Consistency and sample sizes for tested ethnicities.}
\begin{tabular}{@{}cccc@{}}
\toprule
Ethnicity & Faces & Region             & Consistency (Fleiss-$\kappa$) \\ \midrule
Ethiopian & 49    & East Africa        & 0.58        \\
Gambian   & 46    & West Africa        & 0.89        \\
Sardinian & 36    & Southern Europe    & 0.49        \\
Icelandic & 27    & Northern Europe    & 0.94        \\
Balochi   & 27    & Western South Asia & 0.35        \\
Bengali   & 47    & Eastern South Asia & 0.7         \\
Korean    & 34    & East Asia          & 0.96        \\
Filipino  & 32    & South East Asia    & 0.47       \\ 
\bottomrule
\end{tabular}
\label{tab:ethstats}
\end{table}

\begin{figure}
  \centering
  \includegraphics[width=\linewidth]{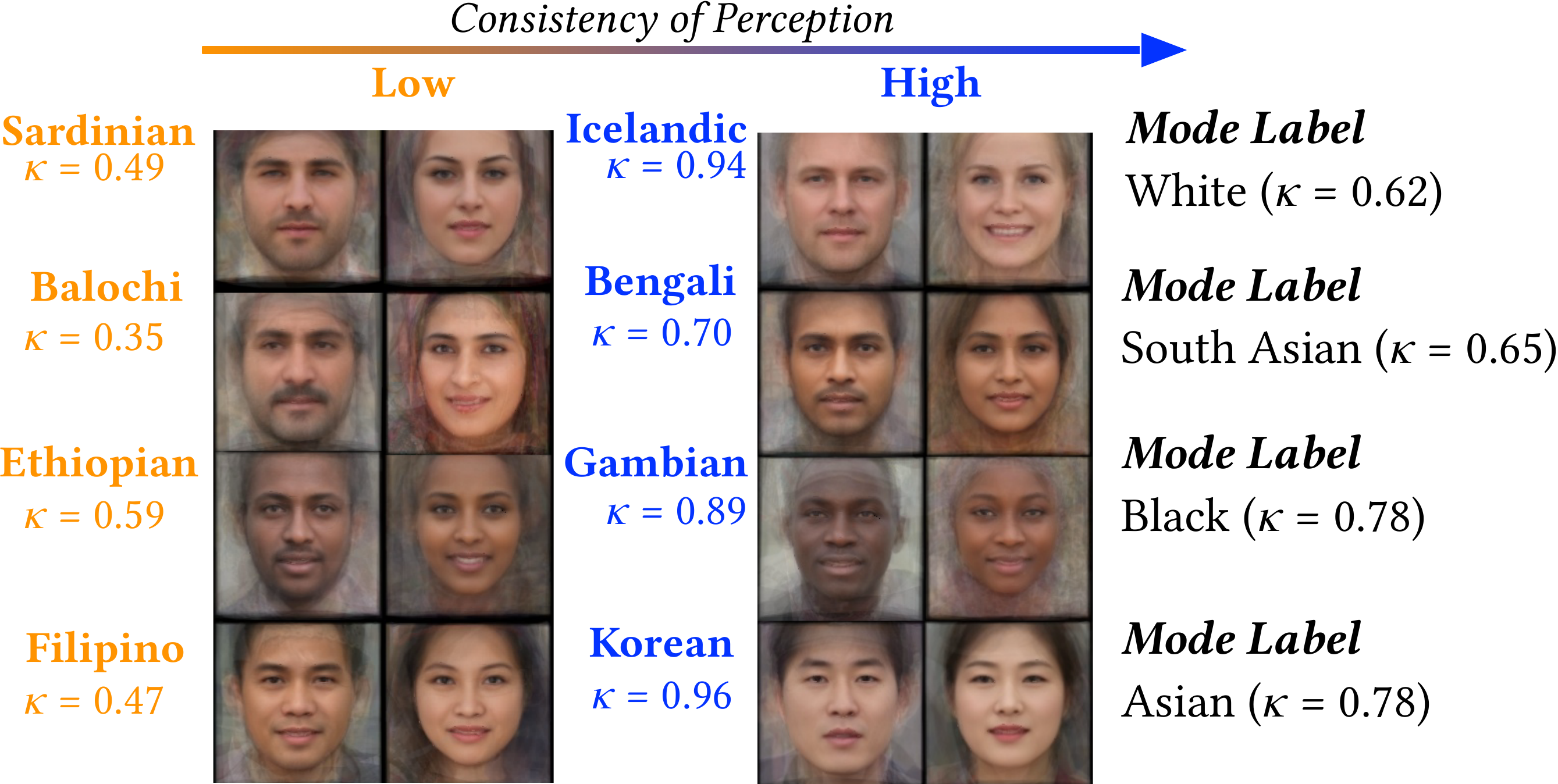}
  \caption{Some ethnicities fit into racial stereotypes of datasets, others don't. Displayed are the average faces of each ethnicity used for experiments, alongside their most frequently predicted racial category and Fleiss-$\kappa$ scores.}
  \Description{The 1907 Franklin Model D roadster.}
\end{figure}
\label{fig:stereotypes}

\section{Discussion}
Our goal was to understand how each dataset constructs racial categories by assigning racial labels to faces, how consistent and similar the constructed categories are across datasets, and the beliefs they encode about who belongs to a racial category, and how clearly they belong to a racial category. 

\subsection{Experimental Results}
The training of the classifier ensemble $\mathcal{C}$ was uneventful. The most surprising conclusion is perhaps that no dataset clearly outperformed the others in its ability to encode racial categories that generalize to others. 
Another interesting result is that the number of identities, or unique individuals, in a dataset is much more important for encoding generalizable racial categories than the sheer number of images.

We then moved onto measuring the self-consistency of the racial assignments in each dataset. 
Intuitively, one would expect the datasets to be self consistent if a classifier is able to learn generalizable concepts from the dataset. Seen in this light, the result that the datasets are self-consistent is not surprising. 
However, it is interesting that despite severely reducing the amount of faces and identities available to each member of a self-ensemble due to splitting the dataset, each self-ensemble members manages to learn concepts of race that are substantially more similar to its cohorts in the self-ensemble than to the concepts of race learned by classifiers trained on other datasets. 
This strongly suggests that the rules annotators use to assign a racial category to a face are relatively simple, and subsequently brittle when it comes to novel faces like ethnicities unrepresented in the training set. 

After self-consistency was established, we undertook a closer examination of whether inconsistency in racial identification is systematic, or largely due to random variations.
If inconsistency in racial identification was largely due to random variations, many of our findings would be uninteresting. However, we found that every racial category contained members who were strongly identified as belonging to that category - they had very few or no images on which the classifiers disagreed about the race of the individual - and we found that every racial category contained members on which the classifiers strongly disagreed by making conflicting racial assignments to their pictures. 
This result is particularly important: it shows that \emph{individuals}, represented in BFW and RFW by a set of pictures depicting the individual, are systematically the subject of racial disagreements. 
In the alternative case, individuals pictures would be the subjects of racial disagreements, with the disagreed-upon and agreed-upon images uniformly spread across individuals. 
Another interesting consequence of this experiment is that classifiers showed relatively high agreement on individuals labeled as Black, compared to the high levels of disagreement seen for individuals labeled as South Asian or White.
There is no intrinsic reason why this should be so, and our hypothesis is that all datasets include and recognize a very specific type of person as Black - a stereotype -  while having more expansive and less consistent definitions for other racial categories. 

To gain more insight into how ethnicity and stereotypes interplay in forming racial categories, we conduct our final experiment, which involves testing our classifier ensemble on the faces of members belonging to specifically chosen ethnic groups. The results are particularly illuminating. 
Despite the small size, we observed a large difference in the consistency of racial perception across ethnic groups. 
It is possible to explain some of the results purely probabilistically - blonde hair is relatively uncommon outside of Northern Europe, so blond hair is a strong signal of being from Northern Europe, and thus, belonging to the White category. 
The others are more difficult to explain. It is not so obvious why the Filipino sample is not consistently seen as Asian while the Korean sample is.
The Ethiopian sample vs the Gambian sample is perhaps the most interesting, and can explain some of the other results we have seen.
Ethiopia is in East Africa, and most of the African diaspora in the United States consist of individuals from West or Central Africa \cite{brycGeneticAncestryAfrican2015}.
If the datasets are biased towards images collected from individuals in the United States, then East Africans may not be included in the datasets, which results in high disagreement on the racial label to assign to Ethiopians relative to the low disagreement on the Black racial category in general.

\subsection{A Clear and Present Danger}
Racial categories and systems are localized in space and time. The meaning of White has shifted over time \cite{saenzInternationalHandbookDemography2015} in the United States. 
An individual who is Black in the United States may not be Black in Brazil \cite{tellesRacialAmbiguityBrazilian2002}. 
A racial system that "makes sense" in the United States may not in other parts of the world. It may not "make sense" a few years into the future. However, algorithms and datasets have the power to shape society and social processes, just as they are shaped by society. 
Already, demographic estimation algorithms using racial categories are being deployed for advertising \cite{olsonQuietGrowthRaceDetection2020}. 
Advertising has the power to shape preferences, desires, and dreams - and it is easy to envision a future in which people are algorithmically offered different dreams and products based on their race. 
Minorities in the United States are already the target of harmful advertising \cite{cnnBillionsSpentAds}, and the potential for discrimination in online advertising \cite{speicherPotentialDiscriminationOnline} has been noted.

Racial categories and systems are fundamentally created by humans, and embedded in a cultural context. 
The encoding of racial categories in datasets, which are then used to train and evaluate algorithms, removes the racial categories from their cultural context, and turns them into concrete, reified categories. 
Datasets commonly build on other datasets (all but one of the datasets studied in this work do so), and so potentially reproduce and amplify the biases of their predecessors.
Divorced from their relativity and cultural context, packaged into neat, concrete disjoint sets, racial categories take on an independent, real existence.
Privileged, wealthy, industrialized countries in which most AI research is conducted have the benefit of being the primary producers of datasets, and package their racial systems into datasets. 
Effectively, this is a potentially dangerous form of cultural export, and eventually, the exported racial systems will take on a validity outside of their cultural context, with potentially dangerous consequences. 

Our experiments with ethnic stereotypes illustrate and underscore the fact that racial categories are not a homogenous mass.
Having demographic equality in the form of racial categories in a dataset does not mean a dataset is diverse, because not all White, Indian, Black, or Asian people are the same. 
A dataset can have equal amounts of individuals across racial categories, but exclude ethnicities or individuals who don't fit into stereotypes.
The rise of generative models and the controversies that have followed as a lack of diversity \cite{LessonsPULSEModel2020} present another example of this effect. An image superresolution model, when provided low resolution pictures of minorities as input, sometimes turned them into White faces.
This was due to imbalances in the underlying training set: it contained mostly White faces. 
However, "fixing" datasets by demographic balancing misses subtleties. Suppose the underlying dataset was demographically balanced with respect to racial categories in the way computer vision datasets are, and we provided it a low resolution picture of an Ethiopian or a Baloch. It may erase their ethnic identity but could produce an image belonging to the "correct" racial category as output. This may be a dubious fix.

\section{Conclusion}
We empirically study the representation of race through racial categories in fair computer vision datasets, and analyze the cross-dataset generalization of these racial categories, as well as their cross-dataset consistency, stereotyping, and self-consistency. 
We find that some categories are more consistently represented than others, but all racial categories transfer across dataset boundaries. 
Furthermore, we find evidence that racial categories rely on stereotypes and exclude ethnicities or individuals who do not fit into those stereotypes. 
We find systematic differences in how strongly individuals in biometric datasets are ascribed a racial category. We point out the dangers and pitfalls of using racial categories in relation to diversity and fairness.

\bibliographystyle{ACM-Reference-Format}
\bibliography{sample-base}


\end{document}